\documentclass{ecai}
\usepackage{graphicx}
\usepackage{latexsym}
\usepackage{verbatim}
\usepackage{hyperref}
\usepackage{caption}
\usepackage{multirow, makecell, tabularx, ragged2e}
\newcolumntype{C}{>{\Centering\arraybackslash}X} 
\newcolumntype{L}{>{\RaggedRight\arraybackslash}X}
\newcolumntype{S}{>{\hsize=.4\hsize}L}

\begin{document}

\begin{frontmatter}

\title{Refining the Responses of LLMs by Themselves}

\author[A]{\fnms{Tianqiang}~\snm{Yan}\thanks{Tianqiang Yan. Email: 222010015@link.cuhk.edu.cn}}
\author[B]{\fnms{Tiansheng}~\snm{Xu}} 

\address[A]{Future Network of Intelligence Institute, The Chinese University of Hong Kong, Shenzhen 518172, P. R. China}
\address[B]{Department of Mathematics, Faculty of Nature Science, Imperial College, London SW7 2AZ, United Kingdom}

\begin{abstract}
In the past few years, Large Language Models (LLMs) have generated unprecedented enthusiasm, with models like GPT providing human-like responses for a wide range of inquiries in nearly all domains. However, these models often fail to deliver satisfactory answers aligned with users’ specific needs on their first attempt, necessitating multiple iterations and additional user input to refine the responses. This can lead to an unnecessary investment of time and effort from users. In this paper, we propose a simple yet efficient approach based on prompt engineering that leverages the large language model itself to optimize its answers without relying on auxiliary models. We introduce an iterative self-evaluating optimization mechanism, with the potential for improved output quality as iterations progress, removing the need for manual intervention. The experiment's findings indicate that utilizing our response refinement framework on the GPT-3.5 model yields results that are on par with, or even surpass, those generated by the cutting-edge GPT-4 model. Detailed implementation strategies and illustrative examples are provided to demonstrate the superiority of our proposed solution.
\end{abstract}

\end{frontmatter}

\section{Introduction}
%
\subsection{Revisiting Large Language Models}
Large Language Models (LLMs) have become a significant development in deep learning techniques, which allows them to understand and generate natural language using vast amounts of textual data. LLMs have shown exceptional potential and flexibility in various Natural Language Processing (NLP) and Natural Language Generation (NLG) tasks, such as text summary, machine translation, sentiment analysis, content creation, and conversational AI. These models are trained using a self-supervised learning approach where they learn from unlabeled data by predicting the following word or token in sequential data. This process enables LLMs to decipher the syntax, semantics, and general knowledge of human language while also retaining significant amounts of factual information retrieved from the training dataset.

The emergence and evolution of LLMs were due to the advancements in transformer models, a type of neural network that utilizes attention mechanisms to encode and decode sequential data. Vaswani et al. first proposed transformer models \cite{Vaswani2017AttentionIA}, and since then, many variations, including BERT \cite{Devlin2019BERTPO}, GPT-3 \cite{Brown2020LanguageMA}, XLNet \cite{Yang2019XLNetGA}, and XLM-RoBERTa \cite{Conneau2019UnsupervisedCR}, have been developed, demonstrating unrivaled performance in several NLP benchmarks and tasks, further highlighting the potency and versatility of LLMs. 

OpenAI’s ChatGPT stands as one of the most renowned LLMs in the field, of which the latest version is based on GPT-4 \cite{openai2023gpt4}. GPT-4 has demonstrated performance that rivals or exceeds human experts in numerous interdisciplinary tasks, while offering support for multimodal data and expanding the range of applications for large language models to a previously unattained scale. At present, the multimodal features of GPT-4’s ChatGPT remain inaccessible to most users. Nonetheless, in regular conversations, the latest version of ChatGPT has exhibited considerable enhancements in understanding and response capabilities compared to its earlier iterations. In addition, the well-developed commercialization of ChatGPT, exemplified by Microsoft’s GPT-4-based Chat with Bing and Office Copilot, as well as a multitude of third-party applications using GPT APIs, has facilitated the gradual permeation of LLM concepts and applications across diverse fields and demographics. This has established a significant milestone in the realm of computer science.

The superiority of LLMs represent a pivotal advancement in NLP research, offering new prospects for language generation, dialogue systems, and creative writing. As LLMs continue to evolve, they are expected to play an increasingly critical role in dictating the direction of natural language processing and machine learning research.

\subsection{Studies on refining the responses of LLMs}
Despite their impressive capabilities, these models are not without limitations: obtaining a user’s desired answer in a single attempt remains a challenging endeavor. Various factors contribute to this issue, such as biases inherent in training data and model architectures, which can result in incorrect or contextually inappropriate responses \cite{Radford2018ImprovingLU}. Moreover, the lack of explainability and transparency in the decision-making process of these black-box models further exacerbates the difficulty in optimizing model outputs for user needs \cite{Bender2021OnTD}. This phenomenon is primarily attributed to the challenges faced by these models in comprehending nuanced and highly specialized contexts or adhering to specific writing styles and formats while generating responses, which often leads to inconsistencies and deviations from the desired output. \cite{Zhang2019DIALOGPTL, Wang2019SuperGLUEAS}.

The Reinforcement Learning with Human Feedback (RLHF) mechanism is a recent advancement in the field of language learning models (LLMs) that aims to optimize their interactive responses to human users. This innovative approach is designed to incorporate human feedback in training LLMs to generate more effective, accurate, and contextually relevant responses, while mitigating potential pitfalls associated with traditional reinforcement-learning-based methods \cite{Brown2020LanguageMA}. One significant advantage of the RLHF mechanism lies in its ability to leverage both expert demonstrations and preference comparisons to build a reward model for the LLM, which enables the model to adapt and improve its response generation based on human-provided feedback \cite{Christiano2017DeepRL}. In order to train the RLHF mechanism to optimize the LLM’s responses to human users, an initial set of demonstrations is provided by human experts who interact with the LLM, generating high-quality responses. These demonstrations are then utilized to perform supervised fine-tuning \cite{Ouyang2022TrainingLM}. Building upon this foundation, the mechanism further incorporates user preference comparisons through a process wherein the LLM generates multiple candidate responses, and users are asked to rank or rate these responses according to their relevance, usefulness, and quality. To adjust and update the reward model accordingly, the RLHF mechanism employs an algorithm that computes the gradients of the reward model based on the aggregated user feedback \cite{Christiano2017DeepRL}.

Despite RLHF’s numerous advantages in training LLMs, one major concern is the possibility of negative side effects from over-optimizing to human feedback, potentially leading the LLM to generate uninformative or excessively verbose responses in order to maximize its perceived reward \cite{Ouyang2022TrainingLM}. Additionally, the reliance on human-generated demonstrations and feedback inherently introduces the potential for bias or inconsistency into the training process, which may influence the performance and behavior of the resulting LLM. In the meantime, the integration of RLHF in LLMs demands high quality volunteer or user responses, leading to increased time and monetary costs.

Another extensively examined research study elucidated that LLMs exhibit a chain-of-thought cognition, and concurrently emphasized that deconstructing the inferential procedure of a problem via chain-of-thought prompting may serve to augment the model’s proficiency in addressing intricate challenges \cite{Wei2022ChainOT}. Fundamentally, this concept necessitates users to meticulously dissect their queries prior to inputting them into comprehensive language models or to encourage the model to deliver responses accompanied by an elaborate reasoning process. If the RLHF optimization procedure is encapsulated into four stages: “User formulates a query, model proffers a resolution, user evaluates the quality of the solution, model fine-tunes based on assessment,” then chain-of-thoughts prompting embodies a more efficacious application-tier feedback optimization technique. This is attributed to its regardlessness of whether the model itself has undergone optimization, and its potential capacity to facilitate the model in producing dependable replies in a singular endeavor. Nonetheless, the challenge with the chain-of-thought theory is that if a user’s query is manually dissected, the model’s ability to generate a reliable response hinges on whether the user has accurately provided the decomposed version of the corresponding question. On the other hand, if the user requests the model to deliver step-by-step answers, but there is a discrepancy between the model’s interpretation of the problem and the user’s original purpose, the ultimate solution can also be entirely off-course. In summary, both RLHF and chain-of-thought prompting offer valuable advantages. As such, our aim is to integrate the strengths of these two methods and develop a fast to deploy, fully automated strategy to improve the performance of LLMs.

In this study, we concentrate on the exploration of application mechanisms employed by prevalent LLMs. Our approach is predicated exclusively upon user inquiries, LLM responses, and judicious supplementary prompts, aiming to enhance the quality of LLM feedback. It’s important to note that the term “quality” used here (including the same expression that appears later) encompasses multiple metrics, including but not limited to the accuracy, comprehensiveness, and conciseness of the answer. Explicitly, we propose a feasible methodology that obviates the necessity for ancillary models or supplementary manual interventions, enabling an LLM to autonomously refine its response to a query through a prompt-driven, adaptively iterative self-assessment and optimization process. With this paradigm as our goal, we design a viable, general-purpose optimization mechanism that is inspired the ideas of conversational reinforcement learning and chain-of-thought. Our contributions can be concluded as:

\begin{itemize}
\item We provide a novel paradigm to refine LLMs' responses in an independent, application-level, and fully automatic way. 

\item Our paradigm, along with the implementation on its basis, allows instant deployment with any available LLM APIs, while requiring nearly zero development knowledge.

\item The implementation is examined with possible daily inquiries, and the joint optimization scheme achieves overall the best outcome.

\end{itemize}

The following content of this report is arranged as follows: In Section II, the proposed optimization scheme is introduced, and the three derived solutions are described. In Section III, the detailed settings and the results of our testing are demonstrated. Eventually, we conclude our study in Section IV. 

\section{The adaptive optimization paradigm}
In this paper, we design a highly-efficient, fully-automatic interaction mechanism grounded in the application layer of large language models (LLMs). Our approach enables adaptive optimization of AI-generated responses without necessitating human involvement, while simultaneously eschewing the need for fine-tuning the underlying language model or introducing supplementary models. Additionally, the framework’s exclusive focus on the application layer makes it remarkably convenient for users to integrate it with LLM APIs. In this section, we will provide a thorough explanation of the optimization process.

\begin{figure}[htbp]
\centering
\includegraphics[width=0.5\textwidth]{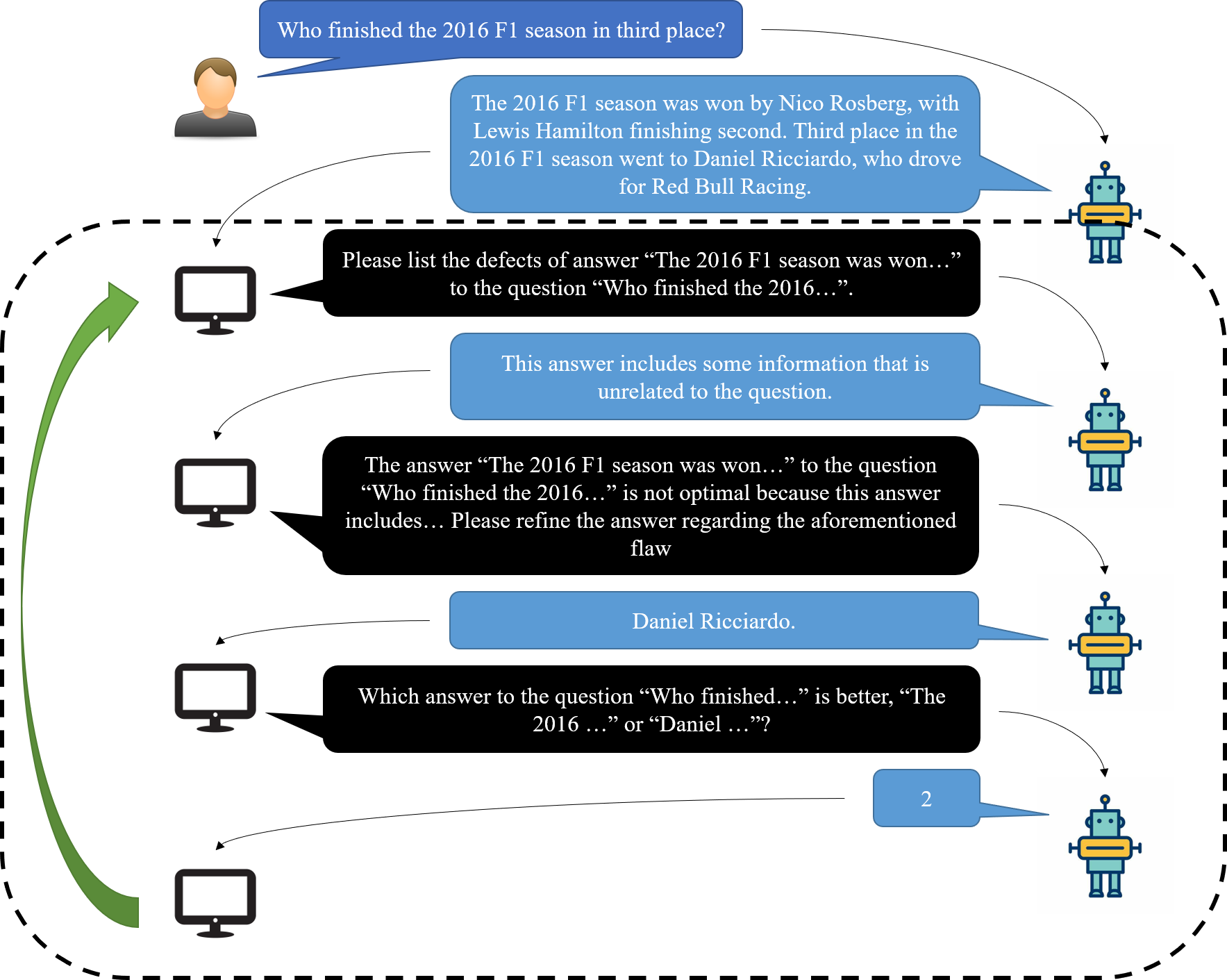}
\captionsetup{justification=centering}
\caption{This is an illustration of using our iterative optimization framework to enhance the answers provided by LLM in response to user queries. The diagram depicts one iteration of the optimization process, involving three agents: the user (depicted as an avatar), a remote LLM server (represented by a robot), and a terminal (symbolized by a computer). The automation loop is enclosed within a dashed box.} 
\label{fig:1}
\end{figure}

\subsection{The overall framework}
The proposed optimization process can be outlined in the following steps:

\begin{enumerate}
\item User step: The user inputs a query into the terminal and sends it via an API to the remote LLM server, with a designated maximum number of optimization iterations.

\item Automatic step 1: The remote LLM server initially responds to the query by producing a model-generated answer and returning it to the terminal.

\item Automatic step 2: The terminal integrates the user query and the model’s previous response to form a prompt, instructing the LLM model to analyze its initial answer, identify any limitations, and provide feedback accordingly.

\item Automatic step 3: The terminal generates an optimization prompt, combining the LLM’s response, the user’s query, and the deficiency analysis, and sends it to the remote server for improvement. The LLM model infers an updated answer and sends it back to the terminal.

\item Automatic step 4: The terminal receives the optimized response and generates a prompt that combines the optimized answer with the previous response and the user’s query, asking the model to determine if the optimized answer is an improvement. The comparison result is then returned to the terminal.

\item Automatic step 5: If the comparison result shows that the optimized answer is better, the terminal utilizes a greedy strategy to repeat the automatic optimization process (from step 3, automatic step 2) until the maximum iterations have been reached. If the result is not improved, the terminal ends the optimization process and returns the previous response.

\end{enumerate}

Figure \ref{fig:1} offers an intuitive portrayal of our response refinement strategy, exemplified through a simulated optimization process. The optimization process as a whole is mainly automated through the terminal interactions with the model API, with the exception of the user setting an upper limit for the number of optimization loops. In practical applications, the specific optimization process is hidden from the user, and they will receive a refined response directly after inputting their question. 

It should be noted that the interaction logic chain throughout the entire enhance process is first-order, meaning that the scheme does not require the LLM to remember the entire previous optimization process. This is done to prevent large token costs and premature depletion of tokens. Furthermore, the optimization process we have designed can ensure the reliability of this first-order optimization mode. The specific reasons for this will be explained in 2.2.

\subsection{Refining the responses of LLMs by themselves}
We have proposed an iterative optimization paradigm that integrates ideas from self-supervised reinforcement learning and chain of thought. 

\begin{figure*}[tb]
\centering
\includegraphics[width=\textwidth]{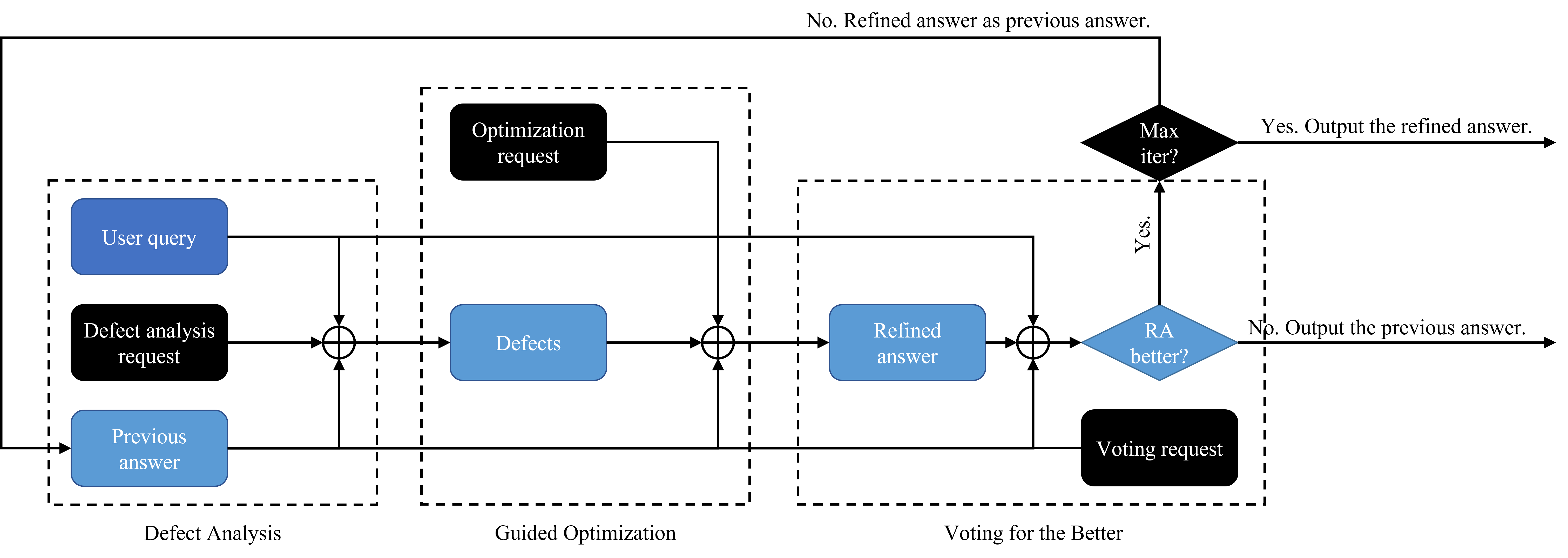}
\captionsetup{justification=centering}
\caption{The flowchart depicting the entire process of an adaptive iterative optimization mechanism. The deep blue rounded rectangles represent variables generated by the user, the black ones represent variables generated by the terminal, and the light blue ones represent variables generated by the remote model. A "\textbf{$\oplus$}" denotes the prompt combination operation.} 
\label{fig:2}
\end{figure*}

\begin{table*}[tb]
        \caption{Settings of the implemented structure of the response optimization}
        \label{table:1}
        \begin{center}
        \begin{tabular}{ccc}
        \hline
        \textbf{LLM} & \textbf{Module} & \textbf{Prompt used} \\
        \hline
        \hline
        \multirow{7}{*}{GPT-3.5-Turbo} & Defect analysis & \makecell[l]{Please list the defects of answer \(a\) to the question \(q\). List the defects in one sentence \\ instead of a list with line breaks!} \\
        \cline{2-3}
        & Guided optimisation & \makecell[l]{The answer \(a\) to the question \(q\) is not optimal because that \(d\). Please refine the answer\\ providing a better one regarding the aforementioned flaw. You should provide nothing\\ but the answer.} \\
        \cline{2-3}
        & Voting for better & \makecell[l]{The question is \(q\), to which there are two optimal answers, one is \(a\), the other one is \(a^*\).\\ Please answer either "1" or "2" if you think one of them is better, or "0" if you think\\ they're equally good. Do not reply anything else than a number!}\\
        \hline
        \end{tabular}
        \end{center}
\end{table*}

\begin{table*}[tb]
        \caption{List of the selected questions and the corresponding reference answers. The reference answers are given by the human expert.}
        \label{table:2}
        \setlength\extrarowheight{2pt}
        \begin{tabularx}{\textwidth}{LL}
        \hline
        \textbf{Question} & \textbf{Reference Answer}\\
        \hline
        \hline
        How to replace the memory on a 2020 Apple M1 processor version MacBook Air? & In fact, the memory of this Macbook is NOT upgradable. \\
        \hline
        How to use the four numbers 2, 2, 8, and 8 along with basic arithmetic operations to obtain 24, with each number used exactly once? & The answer varies. Any answer that meets the requirements is acceptable. \\
        \hline
        The first five numbers in a sequence are 2, 3, 6, 15, and 45, respectively. If the sixth number has only one decimal place and the sequence is incremented, what may be the sixth number? & The answer can be 157.5, or more rigorously, multiple since this is an open question with limited conditions given. \\
        \hline
        Who was the father of Shinkansen? & Shinji Sogō is credited with the creation of the first "bullet train", the Tōkaidō Shinkansen. \\
        \hline
        Why have Formula 1 racing cars adopted the design of halo since 2016? & F1 officially adopted the design in 2018, NOT 2016. It is for protecting the drivers from potential head damage. \\
        \hline
        \end{tabularx}
\end{table*}

\begin{figure*}[tb]
\centering
\includegraphics[width=\textwidth]{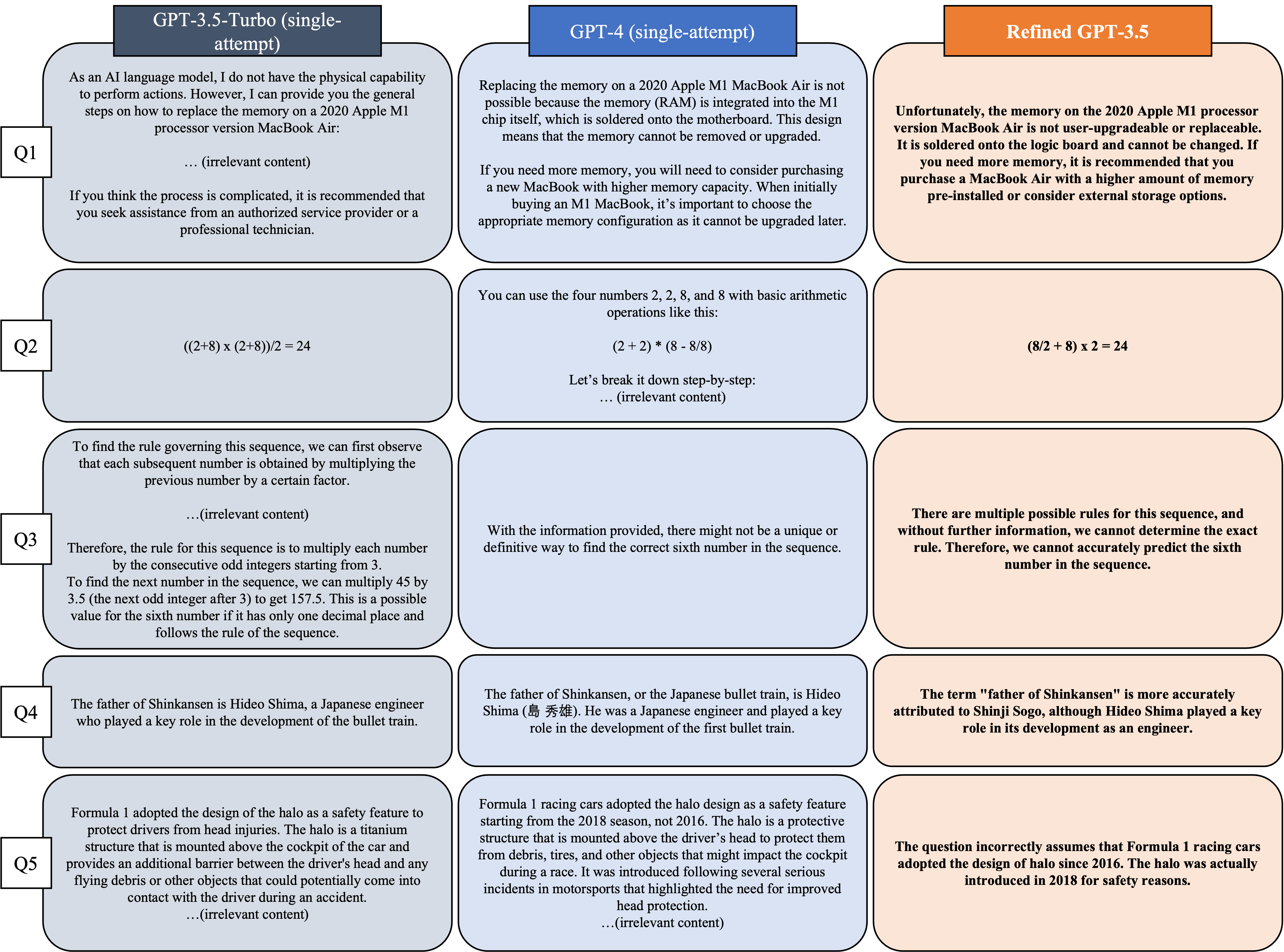}
\captionsetup{justification=centering}
\caption{Answers of the five selected questions generated by the original GPT-3.5-Turbo, the original GPT-4, and the refined GPT-3.5 (horizontal comparison)} 
\label{fig:3}
\end{figure*}

\begin{figure}[htbp]
\centering
\includegraphics[width=0.5\textwidth]{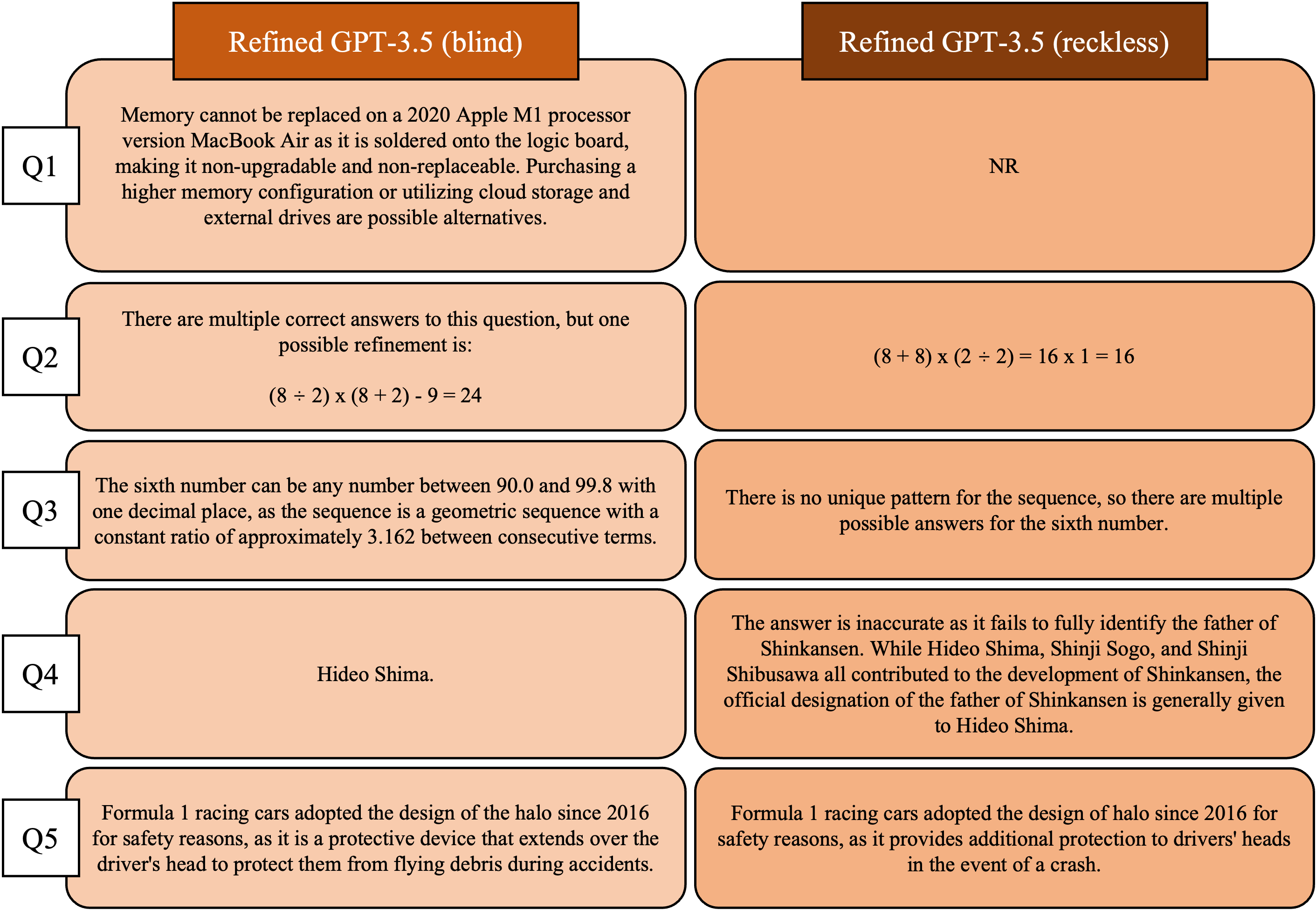}
\captionsetup{justification=centering}
\caption{Answers of the five selected questions generated by the two pruned variations of the refined GPT-3.5, i.e. the \textit{blind refinement} version and the \textit{reckless refinement} version (longitudinal comparison). “NR” refers to that the process is essentially equivalent to the complete refined GPT-3.5 framework, so there’s no need for a duplicate experiment.} 
\label{fig:4}
\end{figure}

Upon revisiting the optimization strategy of RLHF, it is apparent that the two key steps in our refinement mechanism – namely, feedback defect analysis and defect-guiding optimization logic – are fundamentally akin to those of RLHF. However, there are notable distinctions between the two. Whereas RLHF is geared towards LLMs still in their development and debugging phase with a focus on “human feedback,” our optimization approach leverages self-evaluation and self-optimization (SESO) through conversational self-interaction processes that predominantly rely on prompt engineering. Prompt engineering represents a milestone in the field of natural language processing that has emerged concomitantly with the rise of large language models. Enabled by the powerful contextual comprehension and reasoning capabilities inherent in contemporary LLMs, prompt engineering allows for the specification of human-readable prompts tailored to task objectives, thus empowering the model to deliver desired outputs. To facilitate an examination of potential deficiencies in an LLM’s reponse to a given query, all that is required is the transmission of a pre-designed prompt to a remote server which is capable of conveying such a request and providing the desired feedback. To construct a proficient prompt, three critical constituents must be integrated: the initial inquiry posed by the user, the present response generated by the LLM in reference to the inquiry, and prompts that steer the model towards accurate comprehension, analysis, and feedback aligned with the desired objectives. Moreover, in an attempt to mitigate the potential inclusion of extraneous information in LLM outputs, prompts can be augmented with limiting cues. For example, the prompt may conclude with a directive such as “provide the analysis result only” to focus the model’s output on specific aspects relevant to the query.

As our optimization strategy relies on an iterative process, an essential element of our solution is an iterative self-termination mechanism based on a voting method utilizing LLM. When the optimization process advances to this stage, the terminal has already cached the user’s question, the pre-optimized LLM response, and the post-optimized response. The essence of this iterative self-termination strategy is to compare the model’s answers before and after optimization based on the user’s question and select the one it considers superior. The reason for such a judgment mechanism, instead of simply iterating optimization until the user’s maximum optimization times are reached, is that we cannot guarantee that the LLM output is 100\% reliable. In simple terms, whether the optimized answer generated by the model truly achieves the goal of “optimization” is not entirely certain, depending on many factors, including whether the model correctly understands the user’s intent, and whether it accurately digests the defects in the previous answer and makes targeted improvements. As the solution operates at the application level, and the model remains a black box, this self-termination strategy supports adaptive iterations while avoiding any potential negative consequences from further optimization beyond the optimal point. To summarize, this iterative optimization self-termination mechanism contributes to the stability of the output while supporting the process’s adaptability. 

This mechanism is also prompt-driven. The terminal integrates the user’s query, the current and previous answers from the model, and then employs a voting prompt to send the combined input to a remote server. A newly initialized model returns its judgment result. To facilitate the terminal in producing appropriate responses based on the voting result, it is critical to add an limiting instruction. Assuming that the labels of the responses before and after optimization are “1” and “2” respectively, the purpose of the limiting instruction to allow the remote model to only return content limited to be one of the two labels. When the model determines that the current response would better answer the user’s query, it employs a greedy strategy by repeating the previously mentioned optimization steps. On the other hand, if the model determines that the previous answer is still the best response, it returns that answer to the user. 

Another crucial design aspect of our approach is that the response optimization mechanism of this process is not blind, which attributes to the defect-guided enhancement. The inspiration for this optimization method comes from our understanding of the concept of the chain of thought \cite{Wei2022ChainOT}. The idea fundamentally involves using explicit guidance or requiring the model to output the reasoning process in order to improve the robustness of the output of large language models as much as possible. The core purpose of this idea is to prevent the model from blindly searching and generating answers. In the application scenario targeted by our solution, skipping the defect analysis and guided optimization steps would lead to a completely random optimization path, which will further result in the instability of the optimization process. For example, assuming the user’s question is “Where were the 2012 Olympics held?” and the model’s initial response is “The 2012 Olympics took place in London, UK, opening on July 27th and closing on August 12th.” It can be challenging to optimize this answer, even for a human. However, by providing some guidance information to the model, such as “The original question only asks about the location of the Olympics, while the previous answer includes irrelevant time information,” the model can focus its optimization effort and remove the unnecessary time information from the refined answer with high probability. 

We can summarize the entire process into a flow structure, as shown in Figure \ref{fig:2}, based on the various optimization nodes mentioned above. The framework operates with first-order memory since it ensures that each iteration produces a results that is theoretically better than all previous optimization results. Thus, the procedure effectively avoids the accumulation of previous outcomes, which could lead to an increase in token consumption as the number of iterations increases. It's noteworthy that since prompts are based on natural language, the design of prompts involved in each module in the figure may vary. Our focus in this paper is to provide such a comprehensive response optimization framework. In Section 3, we present a series of experiment results to demonstrate the effectiveness of our scheme. These experiments  showcase the superiority of our approach in refining the responses generated by large language models.

\section{Testing and result analysis}
\subsection{The implementation and experiment design}
We have publicly released the source code of an intelligent conversational programme on GitHub, implemented based on the reponse refinement paradigm discussed in this article, while following a flexible modular design as Figure \ref{fig:2}. The project is available through the following link: \texttt{https://github.com/henryyantq/OptimaLLM}. Here, we provide details of the model configurations and prompts used, as listed in Table \ref{table:1}. 

To date, public APIs for other large language models are not yet available. Thus, in this section, we will only use OpenAI's GPT as the target model for our optimization framework. As demonstrated in Table \ref{table:1}, the scheme is applied to the GPT-3.5-Turbo model (hereafter referred to as the \textit{refined} GPT-3.5), which presents a promising opportunity for improvement due to its status as an earlier version of the GPT family offering the chat completion interface. Additionally, the GPT-3.5-Turbo boasts advantages in terms of faster response generation and smaller computational overheads, making it a feasible choice even if multiple iterations are required to improve the quality of the original model’s responses. The consumption of computational resources and time is fully manageable under these circumstances.  

\begin{table*}[tb]
        \captionsetup{justification=centering}
        \caption{The comprehensive assessment results provided by the human expert. The model name indicated in the corresponding row and column refers to the model whose generated answer is acknowledged as having the best performance on that specific question and quota. The evaluation targets only include the three models (GPT-3.5-Turbo, GPT-4, and refined GPT-3.5) in Figure \ref{fig:3}.}
        \label{table:3}
        \setlength\extrarowheight{2pt}
        \begin{tabularx}{\textwidth}{LLLLLL}
        \hline
        \textbf{Quota} & \textbf{Q1} & \textbf{Q2} & \textbf{Q3} & \textbf{Q4} & \textbf{Q5}\\
        \hline
        \hline
        Accuracy & \textbf{refined GPT-3.5} = GPT-4 & \textbf{refined GPT-3.5} & All & \textbf{refined GPT-3.5} & \textbf{refined GPT-3.5} = GPT-4\\
        \hline
        Conciseness & \textbf{refined GPT-3.5} & \textbf{refined GPT-3.5} & hard to decide & \textbf{refined GPT-3.5} & \textbf{refined GPT-3.5}\\
        \hline
        Completeness & GPT-4 & \textbf{refined GPT-3.5} & GPT-3.5-Turbo & \textbf{refined GPT-3.5} & GPT-4\\
        \hline
        \end{tabularx}
\end{table*}

We have chosen five problems commonly faced in daily human life to evaluate the model, as presented in Table \ref{table:2}. Out of the five questions, questions 1, 4, and 5 are factual and questions 2 and 3 are inferential. Of the three factual questions, questions 1 and 5 themselves are somewhat misleading, particularly question 5, which requires the model to identify erroneous information within the question. As for the inferential questions, questions 2 and 3 have multiple correct answers. Therefore, the model is considered correct if it recognizes and provides any or all of the solutions that meet the criteria, with question 2 requiring the model to provide at least one accurate answer. 

The entire experiment procedure consists of two main stages. In the first stage (the horizontal comparison testing), we compare the responses provided by the primary GPT-3.5-Turbo, the refined GPT-3.5 and the GPT-4 when addressing identical questions. The GPT-4’s response process will remain unoptimized, enabling us to assess whether an earlier LLM with our enhancement process applied can rival its original self as well as the current industry-leading model in terms of response quality. Besides, we compared the computational overhead incurred by the native GPT-4 model and the refined GPT-3.5 model when addressing the same questions, emphasizing the cost benefits of applying our peripheral optimization scheme over using a more advanced model for refined feedback. In the second stage (the longitudinal comparison testing), we carry out a comparative assessment of the influence of the optimization framework’s completeness on the resulting response quality, using the same set of questions. In specific terms, we design two simplified versions based on the original optimization framework. The first simplified version is called “\textit{blind} refinement”, which simplifies the original guided optimization mechanism by allowing the model to optimize without referencing any prompts in each loop. The second simplified version is named “\textit{reckless} refinement”, whereby the voting mechanism is removed, ensuring that the optimization process never automatically stops before reaching the predetermined maximum number of iterations. The goal of this stage is to observe and analyze the influence of various modules within the framework on optimization outcomes, underscoring the significance of module integration.

It’s crucial to elaborate on the measuring approach we have taken. To begin with, all the questions we selected for comparative testing have correct answers, which minimizes the uncertainty of response output, reduces the subjectivity of the evaluation process, and ensures the reliability of the results presented. In addition, since humans still possess a superior understanding and evaluative ability in language and question-answering, the final evaluation of response quality for each question from different models is carried out by a human expert. The assessment of response quality is based on three criteria: accuracy, conciseness, and completeness. Accuracy measures how correct the responses are. Conciseness refers to whether the model’s answers contain a significant amount of unnecessary information. Completeness measures whether the model can address all the key points raised in the question. The human expert takes all three aspects into consideration and provides a comprehensive evaluation accordingly.

Overall, the aforementioned experiments can intuitively illustrates the comprehensiveness of our response optimization process, and highlights its potential for enhancing real-world large language models as well as competing against the very best.

\subsection{Experiment results and analysis}
The outcomes from both phase 1 horizontal comparative testing and phase 2 longitudinal comparative testing are presented in Figure \ref{fig:3} and Figure \ref{fig:4} respectively, meanwhile Table 5 presents a comprehensive assessment of all the response outcomes to each question, as evaluated by the human expert.

To facilitate a more intuitive comparison for readers, we have not included the intermediate results generated during the iterative optimization process. If it is necessary to refer to the intermediate results during subsequent analysis, we will enumerate them at the corresponding location. 

Based on the comprehensive assessment results of the horizontal comparison test (see Table \ref{table:3}), our refined GPT-3.5 model achieved an astonishing 100\% accuracy in answering these questions, a clear advantage over both the original GPT-4 and GPT-3.5-Turbo models without any response optimization strategies. In addition, the refined GPT-3.5 also surpassed all competitors regarding the answer conciseness. Yet in terms of answer completeness, although GPT-4 answered one question incorrectly, our refined GPT-3.5 model could not surpass its advantage in delivering more thorough explanations. Notably, the answers obtained from the refined GPT-3.5 are results of the iterative optimization of the response refinement framework proposed in this paper, built upon the initial answers provided by the native GPT-3.5-Turbo. As it is obvious that GPT-3.5-Turbo performed the worst on account of the results, this demonstrates that our paradigm is capable of refining the accuracy and redundancy of the initial answers to produce new ones with higher quality. Furthermore, although refined GPT-3.5 may require more time for iteration, but its total computational cost is much lower than that of GPT-4. Even when using the same number of tokens, refined GPT-3.5 with a maximum iteration limit of 3 can save 5 to 10 times the API usage consumption, thus greatly reducing the economic burden on users. The reason for the lower computational cost is not only due to the use of a more economical model, but also because of the contribution of the first-order memory of the optimization framework mentioned earlier,since such operation mode does not exponentially accumulate the number of tokens, and the resource cost of each iteration can be considered almost constant.

The organic combination of each module in the optimization framework is also essential for obtaining high-quality responses. From the answers provided by variants of refined GPT-3.5 that have removed one key module for each question (as shown in Figure \ref{fig:4}), it can be seen that both weakening the purpose of optimization (guided optimization mechanism) and removing the ability of self-review and self-stop (voting mechanism) significantly reduce optimization capabilities and even result in negative optimization. 

We found some pivotal clues based on the intermediate results to explain the reasons for the shortcomings. As an example of the iterative optimization process for the second question using the blind refinement framework, the initial response given by the native GPT-3.5-Turbo contained numerous incorrect steps and an incorrect equation: "$(8\times8)+2\div2-9=24$". This answer is problematic because the original question does not include the number 9, does not allow for any numbers other than 8, 8, 2, 2 to be used in the calculation, and the equation on both sides of the equals sign is actually not equivalent. After the first round of iteration, the blind refinement mechanism removed only the cumbersome steps, keeping the erroneous equation as the new response, which was still clearly incorrect. Even after the second and third rounds of iteration, this incorrect equation remained a major part of the answer, resulting in a final optimized solution that was still incorrect. This suggests that allowing the model to optimize without a clear purpose can result in the model repeatedly rewriting the answer without actually improving it. At the same time, in the example of the reckless refinement framework that removes the voting module to deal with the same question, the model provided a correct answer by the end of the second iteration, but unexpectedly produced an incorrect answer by the third iteration. The lack of self-review and self-stop mechanisms means that the model continued to optimize beyond the point where it had already obtained the correct answer, and as a result, ended up outputting an incorrect answer that was generated in the final round of optimization. Both of these typical examples serve to underscore the integral roles played by these two modules throughout the optimization process.

To sum up, We have substantiated through the conclusions of the above experiments that using the iterative optimization framework proposed in this paper on a large language model can significantly enhance the quality of the generated answers at a lower cost Furthermore, it highlights the contributions made by the guided optimization mechanism and the voting mechanism in improving optimization capabilities.

\section{Conclusion}
In this paper, we introduce a fully-autonomous adaptive iterative response optimization paradigm, inspired by concepts from RLHF and the chain of thought. This approach relies solely on simple prompt engineering and the LLM API, without the need for manual intervention, auxiliary models, or access to internal structures and parameters of language models. Specifically, we present a detailed optimization framework utilizing an efficient modular design, applied to the GPT-3.5-Turbo. Our experiments show that our optimization mechanism enables a less-capable model to achieve response quality better than its original self, and even on par with one of the best current models while reducing resource consumption. Through this scheme, we demonstrate that in many situations, the existing question-answering interaction paradigm may not fully harness the potential of generative language models. Appropriately designing prompts and planning response interaction logic is a crucial approach to further unleash a model’s potential. 

\bibliography{ecai}
\end{document}